\documentclass[11pt,letterpaper]{article}
\usepackage[draft]{hyperref}
\usepackage{emnlp2017}
\usepackage{times}
\usepackage{latexsym}
\usepackage{verbatim}
\usepackage{booktabs}
\usepackage{float}
\usepackage{comment}
\usepackage{color}
\usepackage{fancyvrb}
\usepackage{multirow}
\usepackage{array}
\usepackage{graphicx}
\usepackage{adjustbox}
\usepackage{siunitx}
\usepackage{enumitem}
\usepackage{subcaption}
\usepackage{amsmath}

\let\oldFootnote\footnote
\newcommand\nextToken\relax

\renewcommand\footnote[1]{%
    \oldFootnote{#1}\futurelet\nextToken\isFootnote}

\newcommand\isFootnote{%
    \ifx\footnote\nextToken\textsuperscript{,}\fi}

\emnlpfinalcopy

\def\emnlppaperid{1022}

\title{``i have a feeling trump will win..................": \\ Forecasting Winners and Losers from User Predictions on Twitter}

  \author{Sandesh Swamy\textmd{,} Alan Ritter \\ Computer Science \& Engineering\\ The Ohio State University \\ Columbus, OH \\ swamy.14@osu.edu, aritter@cse.ohio-state.edu \And Marie-Catherine de Marneffe \\ Department of Linguistics\\ The Ohio State University \\ Columbus, OH \\ mcdm@ling.ohio-state.edu}
\date{}

\begin{document}

\maketitle
\begin{abstract}
  Social media users often make explicit predictions about upcoming events. Such statements vary in the degree of certainty the author expresses toward the outcome: ``Leonardo DiCaprio will win Best Actor" vs.\ ``Leonardo DiCaprio may win" or ``No way Leonardo wins!". Can popular beliefs on social media predict who will win?  To answer this question, we build a corpus of tweets annotated for veridicality on which we train a log-linear classifier that detects positive veridicality with high precision.\footnote{The code and data can be found at \url{https://github.com/SandeshS/Twitter-Veridicality}}  We then forecast uncertain outcomes using the {\em wisdom of crowds}, by aggregating users' explicit predictions. Our method for forecasting winners is fully automated, relying only on a set of contenders as input.  
  It requires no training data of past outcomes and outperforms sentiment and tweet volume baselines on a broad range of contest prediction tasks.
  We further demonstrate how our approach can be used to measure the reliability of individual accounts' predictions and retrospectively identify surprise outcomes.
\end{abstract}


\section{Introduction}

In the digital era we live in, millions of people broadcast their thoughts and opinions online.
These include predictions about upcoming events of yet unknown outcomes, such as the Oscars or election results.  
Such statements vary in the extent to which their authors intend to convey the event will happen. For instance, (a) in Table \ref{table:examples}  strongly asserts the win of Natalie Portman over Meryl Streep, whereas (b) imbues the claim with uncertainty. In contrast, (c) does not say anything about the likelihood of Natalie Portman winning (although it clearly indicates the author would like her to win).

\begin{table}
    \small
    \centering
    \begin{tabular}{cp{2.5in}}
    \toprule
    (a) & {\it Natalie Portman is gonna beat out Meryl Streep for best actress} \\
    \midrule
    (b) & {\it La La Land doesn't have lead actress and actor guaranteed. Natalie Portman will probably (and should) get best actress} \\
    \midrule
    (c) & {\it Adored \#LALALAND but it's \#NataliePortman who deserves the best actress \#oscar \#OscarNoms $>$ superb acting} \\
    \bottomrule
    \end{tabular}
    \caption{Examples of tweets expressing varying degrees of veridicality toward Natalie Portman winning an Oscar.}
    \label{table:examples}
\end{table}

Prior work has made predictions about contests such as NFL games \cite{sinha-13} and elections using tweet volumes \cite{GermanElections} or sentiment analysis \cite{o2010tweets,Shi12}. 
Many such indirect signals have been shown useful for prediction, however their utility varies across domains.  In this paper we explore whether the ``wisdom of crowds" \cite{surowiecki2005wisdom}, as measured by users' explicit predictions, can predict outcomes of future events.  We show how it is possible to accurately forecast winners, by aggregating many individual predictions that assert an outcome.  Our approach requires no historical data about outcomes for training and can directly be adapted to a broad range of contests.

To extract users' predictions from text, we present TwiVer, a system that classifies veridicality toward future contests with uncertain outcomes.
Given a list of contenders competing in a contest (e.g., Academy Award for Best Actor), we use TwiVer to count how many tweets explicitly assert the win of each contender.  We find that aggregating veridicality in this way provides an accurate signal for predicting outcomes of future contests.  Furthermore, TwiVer allows us to perform a number of novel qualitative analyses including retrospective detection of {\em surprise outcomes} that were not expected according to popular belief (Section~\ref{surprise}).  We also show how TwiVer can be used to measure the number of correct and incorrect predictions made by individual accounts.  This provides an intuitive measurement of the reliability of an information source (Section~\ref{assessing}).






\section{Related Work}

In this section we summarize related work on text-driven forecasting and computational models of veridicality.

{\em Text-driven forecasting models} \cite{smith2010text} predict future response variables using text written in the present: e.g., forecasting films' box-office revenues using critics' reviews \cite{joshi2010movie}, predicting citation counts of scientific articles \cite{yogatama2011predicting} and success of literary works \cite{DBLP:conf/emnlp/AshokFC13}, forecasting economic indicators using query logs \cite{choi2012predicting}, improving influenza forecasts using Twitter data \cite{paul2014twitter}, predicting betrayal in online strategy games \cite{niculae-EtAl:2015:ACL-IJCNLP} and predicting changes to a knowledge-graph based on events mentioned in text \cite{konovalov17}.  These methods typically require historical data for fitting model parameters, and may be sensitive to issues such as concept drift \cite{fung2014google}.  In contrast, our approach does not rely on historical data for training; instead we forecast outcomes of future events by directly extracting users' explicit predictions from text.

Prior work has also demonstrated that user sentiment online directly correlates with various real-world time series, including polling data \cite{o2010tweets} and movie revenues \cite{mishne2006predicting}.  In this paper, we empirically demonstrate that veridicality can often be more predictive than sentiment (Section~\ref{prediction}).

Also related is prior work on {\em detecting veridicality} \cite{de2012did,sogaard2015using} and sarcasm \cite{gonzalez2011identifying}. Soni et al.\ \shortcite{soni2014modeling} investigate how journalists frame quoted content on Twitter using predicates such as {\em think}, {\em claim} or {\em admit}.  In contrast, our system TwiVer, focuses on the author's belief toward a claim and direct predictions of future events as opposed to quoted content.

Our approach, which aggregates predictions extracted from user-generated text is related to prior work that leverages explicit, positive veridicality, statements to make inferences about users' demographics.  For example, Coppersmith et al.\ \shortcite{coppersmith2014measuring,coppersmith2015adhd} exploit users' self-reported statements of diagnosis on Twitter.

\section{Measuring the Veridicality of Users' Predictions}


The first step of our approach is to extract statements that make explicit predictions about unknown outcomes of future events. We focus specifically on {\em contests} which we define as events planned to occur on a specific date, where a number of {\em contenders} compete and a single {\em winner} is chosen.  For example, Table~\ref{tab:oscar_actor} shows the contenders for Best Actor in 2016, highlighting the winner.



\begin{table}[htbp]
    \small
    \centering
    \begin{tabular}{l|l}
    \toprule
    Actor & Movie \\
    \midrule
    \textbf{Leonardo DiCaprio} & \textbf{The Revenant} \\
    Bryan Cranston & Trumbo \\
    Matt Damon & The Martian \\
    Michael Fassbender & Steve Jobs\\
    Eddie Redmayne & The Danish Girl\\
    \bottomrule
    \end{tabular}
    \caption{Oscar nominations for Best Actor 2016.}
    \label{tab:oscar_actor}
\end{table}

\begin{figure*}
    \small
    \centering
    \includegraphics[scale=0.3]{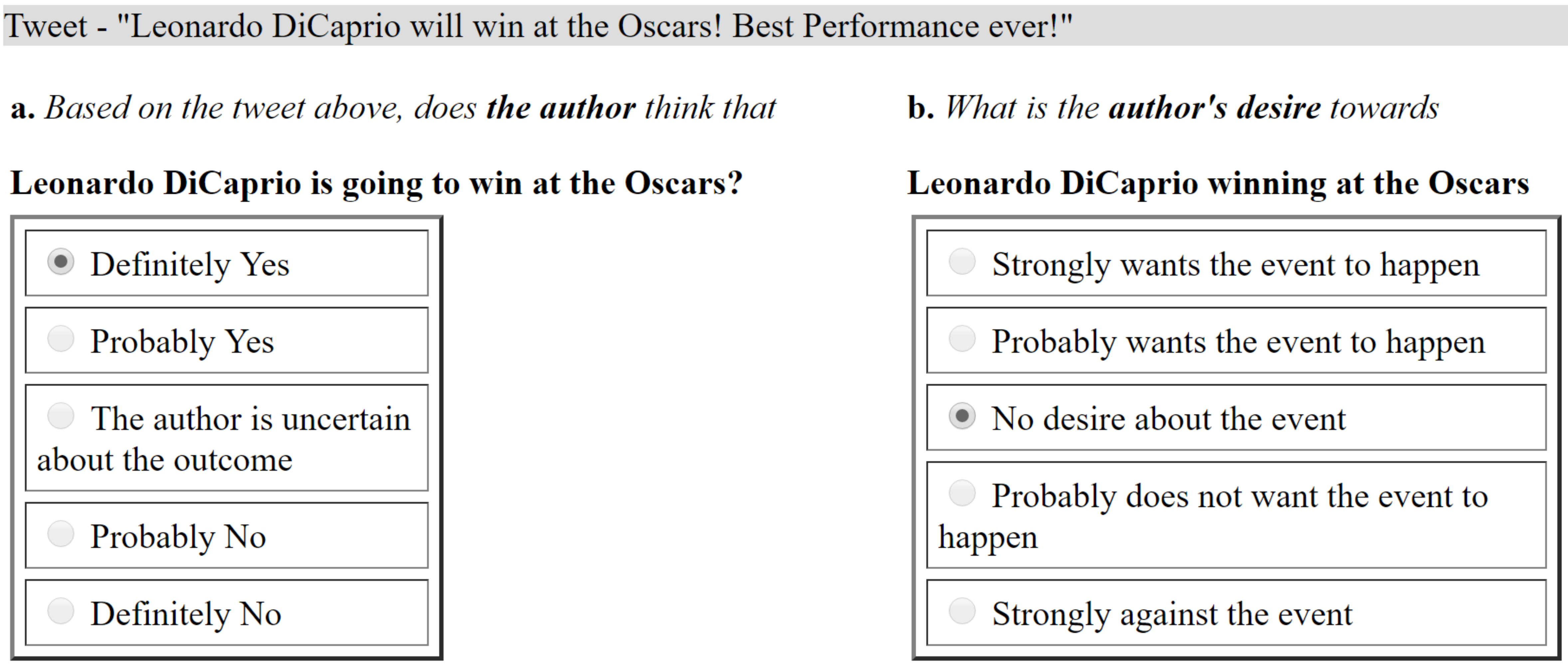}
    \caption{Example of one item to be annotated, as displayed to the Turkers.}
    \label{fig:mturk}
\end{figure*}


To explore the accuracy of user predictions in social media, we gathered a corpus of tweets that mention events belonging to one of the 10 types listed in Table \ref{table:tweet numbers}.  Relevant messages were collected by formulating queries to the Twitter search interface that include the name of a contender for a given contest in conjunction with the keyword \emph{win}.
We restricted the time range of the queries to retrieve only messages written before the time of the contest to ensure that outcomes were unknown when the tweets were written. We include 10 days of data before the event for the presidential primaries and the final presidential elections, 7 days for the Oscars, Ballon d'Or and Indian general elections, and the period between the semi-finals and the finals for the sporting events.  
Table \ref{table:queries}
shows several example queries to the Twitter search interface which were used to gather data. We automatically generated queries, using templates, for events
scraped from various websites: 483 queries were generated for the presidential primaries based on events scraped from ballotpedia\footnote{\url{https://ballotpedia.org/Main_Page}} 
, 176 queries were generated for the Oscars,\footnote{\url{https://en.wikipedia.org/wiki/Academy_Awards}} 18 for Ballon d'Or,\footnote{\url{https://en.wikipedia.org/wiki/Ballon_d\%27Or}} 162 for the Eurovision contest,\footnote{\url{https://en.wikipedia.org/wiki/Eurovision_Song_Contest}} 52 for Tennis Grand Slams,\footnote{\url{https://en.wikipedia.org/wiki/Grand_Slam_(tennis)}} 6 for the Rugby World Cup,\footnote{\url{https://en.wikipedia.org/wiki/Rugby_World_Cup}} 18 for the Cricket World Cup,\footnote{\url{https://en.wikipedia.org/wiki/Cricket_World_Cup}} 12 for the Football World Cup,\footnote{\url{https://en.wikipedia.org/wiki/FIFA_World_Cup}} 76 for the 2016 US presidential elections,\footnote{\url{https://en.wikipedia.org/wiki/United_States_presidential_election,_2016}} and 68 queries for the 2014 Indian general elections.\footnote{\url{https://en.wikipedia.org/wiki/Indian_general_election,_2014}}


We added an event prefix (e.g., ``Oscars" or the state for presidential primaries), a keyword (``win"), and the relevant date range for the event. For example, ``Oscars Leonardo DiCaprio win since:2016-2-22 until:2016-2-28" would be the query generated for the first entry in Table \ref{tab:oscar_actor}.

\begin{table}[h!]
    \small
    \centering
    \begin{tabular}{c}
    \toprule
    Minnesota Rubio win since:2016-2-18 until:2016-3-1\\
    \midrule
    Vermont Sanders win since:2016-2-18 until:2016-3-1\\
    \midrule
    Oscars Sandra Bullock win since:2010-3-1 until:2010-3-7\\
    \midrule
    Oscars Spotlight win since:2016-2-22 until:2016-2-28\\
    \bottomrule
    \end{tabular}
    \caption{Examples of queries to extract tweets.}
    \label{table:queries}
\end{table}

We restricted the data to English tweets only, as tagged by \textit{langid.py} \cite{lui_langid}. Jaccard similarity was computed between messages to identify and remove duplicates.\footnote{A threshold of 0.7 was used.} We removed URLs and preserved only tweets that mention contenders in the text. This automatic post-processing left us with 57,711 tweets for all winners and 55,558 tweets for losers (contenders who did not win) across all events. Table~\ref{table:tweet numbers} gives the data distribution across event categories.

\begin{table}[htbp]
\small
\centering
   \begin{tabular}{lrr}
   \toprule
   Event & \multicolumn{2}{c}{Number of tweets}\\
     & Winners & Losers  \\
    \midrule 
   2016 US Presidential primaries & 20,347 & 17,873\\
   Oscars (2009 -- 2016) & 1,498 & 872\\
   Tennis Grand Slams (2011 -- 2016) & 10,785 & 19,745 \\
   Ballon d'Or Award (2010 -- 2016) & 3,492 & 3,285 \\
   Eurovision (2010 -- 2016) & 261 & 1,421 \\
   2016 US Presidential elections & 9,679 & 3,966 \\
   2014 Indian general elections & 920 & 736 \\
   Rugby World Cup (2010 -- 2016) & 272 & 379 \\
   Football World Cup (2010 -- 2016) & 8,129 & 5,489 \\
   Cricket World Cup (2010 -- 2016) & 2,328 & 1,792 \\
   \bottomrule
\end{tabular}
\caption{Number of tweets for each event category.}
\label{table:tweet numbers}
\end{table}

\begin{figure*}
    \begin{subfigure}{0.24\textwidth}
    \centering
    \includegraphics[width=0.9\linewidth]{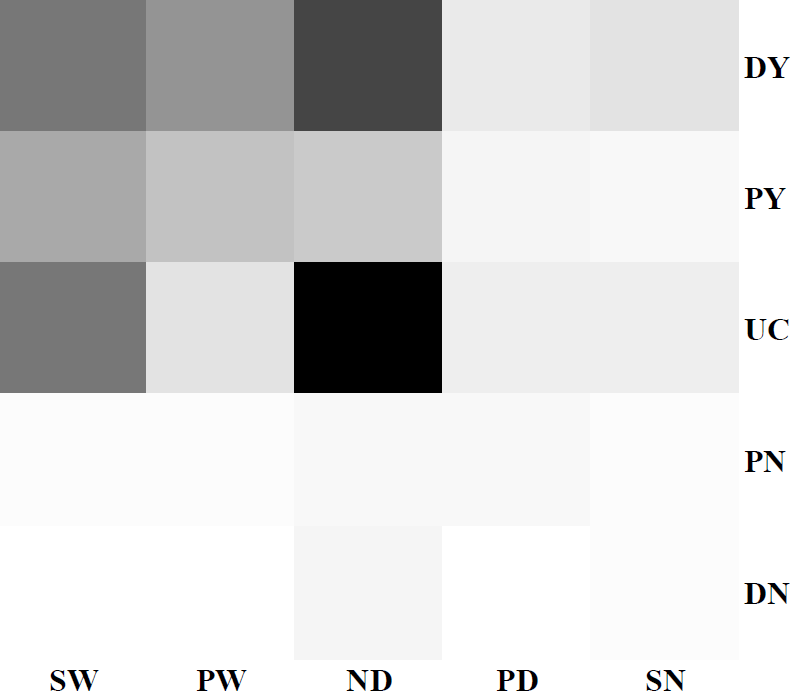} 
    \caption{Oscar winners}
    \label{fig:heat_oscar_win}
    \end{subfigure}
    \begin{subfigure}{0.24\textwidth}
    \centering
    \includegraphics[width=0.9\linewidth]{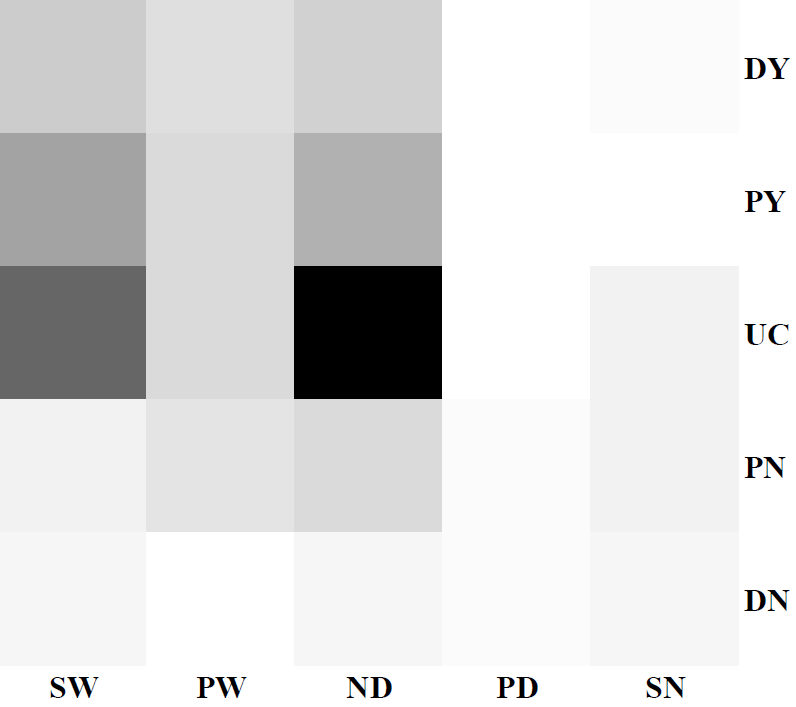}
    \caption{Oscar losers}
    \label{fig:heat_oscar_lose}
    \end{subfigure}
    \quad
    \begin{subfigure}{0.24\textwidth}
    \centering
    \includegraphics[width=0.9\linewidth]{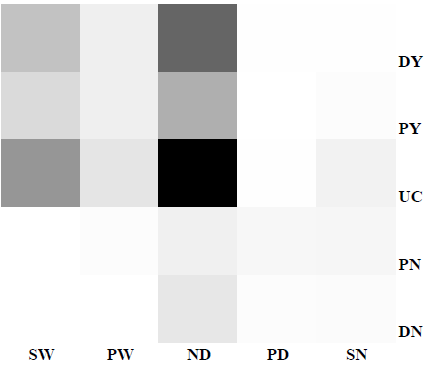} 
    \caption{All events winners}
    \label{fig:heat_all_win}
    \end{subfigure}
    \begin{subfigure}{0.24\textwidth}
    \centering
    \includegraphics[width=0.9\linewidth]{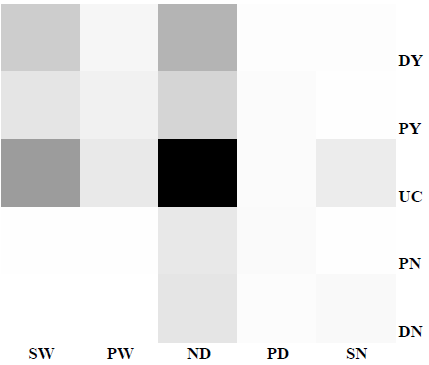}
    \caption{All events losers}
    \label{fig:heat_all_lose}
    \end{subfigure}
    \caption{Heatmaps showing annotation distributions for one of the events - the Oscars and all event types, separating winners from losers. Vertical labels indicate  veridicality (DY ``Definitely Yes", PY ``Probably Yes", UC ``Uncertain about the outcome", PN ``Probably No" and DN ``Definitely No"). Horizontal labels indicate desire (SW ``Strongly wants the event to happen", PW ``Probably wants the event to happen", ND ``No desire about the event outcome", PD ``Probably does not want the event to happen", SN ``Strongly against the event happening").  More data in the upper left hand corner indicates there are more tweets with positive veridicality and desire.}
    \label{fig:heat_all}
\end{figure*}

\subsection{Mechanical Turk Annotation}
We obtained veridicality annotations on a sample of the data using Amazon Mechanical Turk. For each tweet, we asked Turkers to judge veridicality toward a candidate winning as expressed in the tweet as well as the author's desire toward the event. For veridicality, we asked Turkers to rate whether the author believes the event will happen on a 1-5 scale (``Definitely Yes", ``Probably Yes", ``Uncertain about the outcome", ``Probably No", ``Definitely No").  We also added a question about the author's desire toward the event to make clear the difference between veridicality and desire.  For example, ``I really want Leonardo to win at the Oscars!" asserts the author's desire toward Leonardo winning, but remains agnostic about the likelihood of this outcome, whereas ``Leonardo DiCaprio will win the Oscars" is predicting with confidence that the event will happen.

Figure~\ref{fig:mturk} shows the annotation interface presented to Turkers. Each HIT contained 10 tweets to be annotated. We gathered annotations for 
$1,841$ tweets for winners
and $1,702$ tweets for losers, giving us a total of $3,543$ tweets.  We paid \$0.30 per HIT.  
The total cost for our dataset was \$1,000. Each tweet was annotated by 7 Turkers.
We used MACE \cite{hovy_mace} to resolve differences between annotators and produce a single gold label for each tweet.

Figures~\ref{fig:heat_oscar_win} and \ref{fig:heat_all_win} show heatmaps of the distribution of annotations for the winners for the Oscars in addition to all categories. In both instances, most of the data is annotated with ``Definitely Yes" and ``Probably Yes" labels for veridicality. Figures~\ref{fig:heat_oscar_lose} and \ref{fig:heat_all_lose} show that the distribution is more diverse for the losers. Such distributions indicate that the veridicality of crowds' statements could indeed be predictive of outcomes.  We provide additional evidence for this hypothesis using automatic veridicality classification on larger datasets in \S \ref{forecasting_contest_outcomes}.

\subsection{Veridicality Classifier}
The goal of our system, TwiVer, is to automate the annotation process by predicting how veridical a tweet is toward a candidate winning a contest: is the candidate deemed to be winning, or is the author uncertain? For the purpose of our experiments, we collapsed the five labels for veridicality into three: positive veridicality (``Definitely Yes" and ``Probably Yes"), neutral (``Uncertain about the outcome") and negative veridicality (``Definitely No" and ``Probably No").

We model the conditional distribution over a tweet's veridicality toward a candidate $c$ winning a contest against a set of opponents, $O$, using a log-linear model:
\[
P(y=v | c, \text{tweet}) \propto \exp\left(\theta_v \cdot f(c, O, \text{tweet})\right)
\]
where $v$ is the veridicality (positive, negative or neutral).

\begin{figure*}
    \small
    \centering
    \includegraphics[scale=0.40]{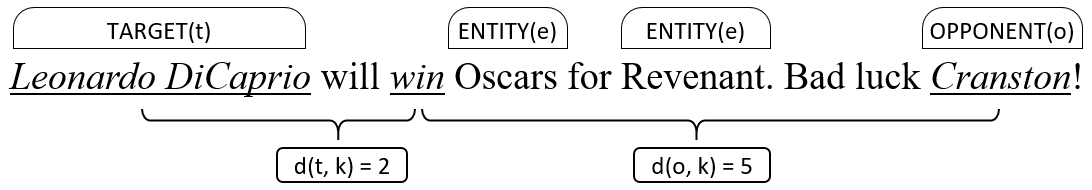}
    \caption{Illustration of the three named entity tags and distance features between entities and keyword \textit{win} for a tweet retrieved by the query ``Oscars Leonardo DiCaprio win since:2016-2-22 until:2016-2-28".}
    \label{fig:feat_extract}
\end{figure*}

To extract features $f(c,O,\text{tweet})$, we first preprocessed tweets retrieved for a specific event to identify named entities, using \cite{Ritter11}'s Twitter NER system.  Candidate ($c$) and opponent entities were identified in the tweet as follows:\\
\noindent - \textsc{target ($t$)}. A target is a named entity that matches a contender name from our queries.\\ 
\noindent - \textsc{opponent ($O$)}. For every event, along with the current \textsc{target} entity, we also keep track of other contenders for the same event. If a named entity in the tweet matches with one of other contenders, it is labeled as opponent.\\ 
\noindent - \textsc{entity ($e$)}: Any named entity which does not match the list of contenders. 

Figure~\ref{fig:feat_extract} illustrates the named entity labeling for a tweet obtained from the query ``Oscars Leonardo DiCaprio win since:2016-2-22 until:2016-2-28". Leonardo DiCaprio is the \textsc{target}, while the named entity tag for Bryan Cranston, one of the losers for the Oscars, is re-tagged as \textsc{opponent}.
These tags provide information about the position of named entities relative to each other, which is used in the features. 

\begin{figure}
    \small
    \centering    
    \includegraphics[scale=0.23]{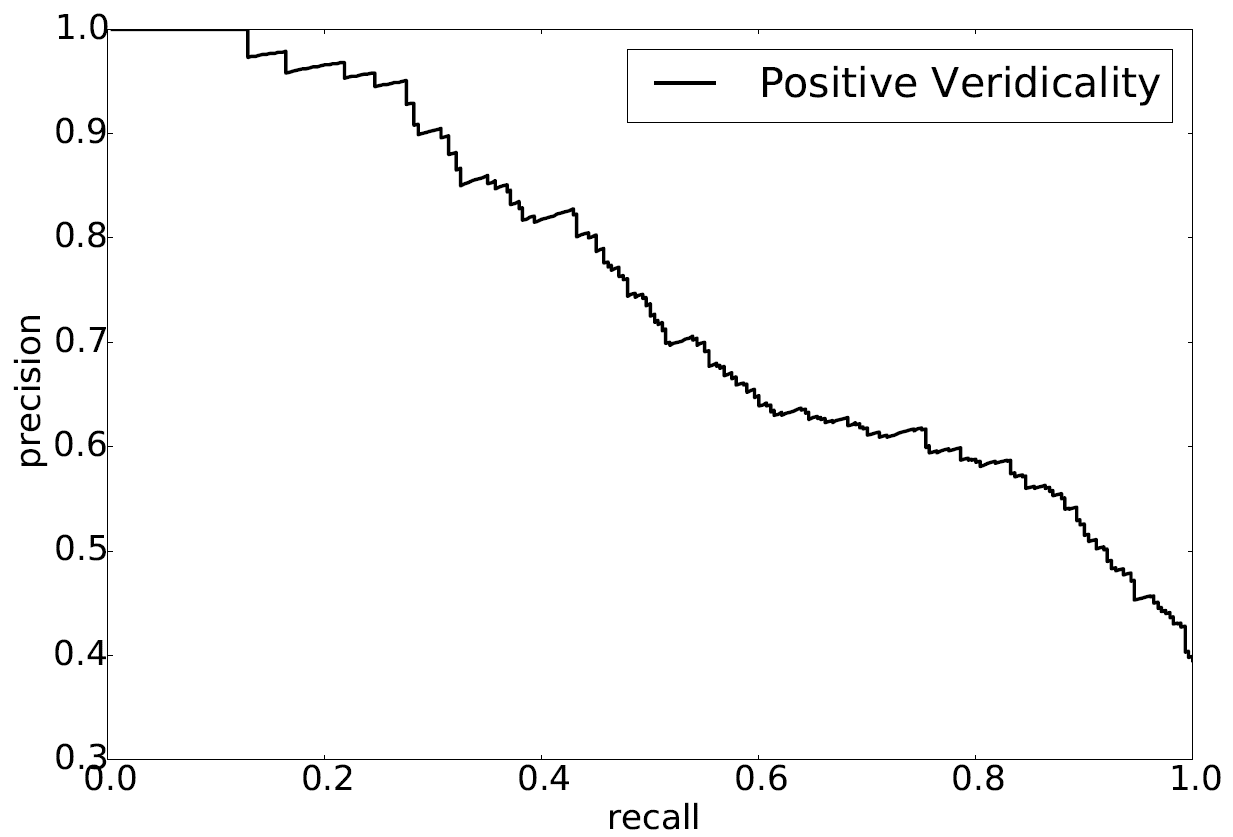}
    \caption{Precision/Recall curve showing TwiVer performance in identifying positive veridicality tweets in the test data.}  
    \label{fig:pos_veridic}
\end{figure}

\subsection{Features}
We use five feature templates: context words, distance between entities, presence of punctuation, dependency paths, and negated keyword.

\begin{table}
    \small
    \centering
    \begin{tabular}{lp{0.8cm}p{0.8cm}p{0.8cm}}
        \toprule
         & P & R & F1\\
         \midrule
         $-$ Context  & 47.7 & \textbf{96.4} & 63.8\\
         $-$ Distance  & 57.5 & 82.5 & 67.7\\
         $-$ Punctuation  & 53.4 & 88.2 & 66.6\\
         $-$ Dependency path  & 56.9 & 85.4 & 68.2 \\
         $-$ Negated keyword  & 56.7 & 86.4 & 68.4 \\
         All features & \textbf{58.7} & 83.1 & \textbf{68.8}\\
         \bottomrule
         & 
    \end{tabular}
    \caption{Feature ablation of the positive veridicality classifier by removing each group of features from the full set.  The point of maximum F1 score is shown in each case.}
    \label{tab:feature_ablation}
\end{table}

\begin{table*}
\small
\centering
    \resizebox{\textwidth}{!}{
    \begin{tabular}{lll|lll}
    \toprule
    \multicolumn{3}{l|}{Positive Veridicality} & \multicolumn{3}{l}{Negative Veridicality}\\
    Feature Type & Feature & Weight & Feature Type & Feature & Weight\\
    \midrule
    Keyword context & \textit{\textsc{target} will \textsc{keyword}} &  0.41 & Negated keyword  & keyword is negated & 0.47\\
    Keyword dep. path & \textit{\textsc{target} $\rightarrow$ to $\rightarrow$ \textsc{keyword}} & 0.38 & Keyword context & \textsc{target} \textit{won't} \textsc{keyword} &  0.41 \\ 
    Keyword dep. path  & \textit{\textsc{target} $\leftarrow$ is $\rightarrow$ going $\rightarrow$ to $\rightarrow$ \textsc{keyword}} & 0.29 & Opponent context & \textit{\textsc{opponent} will win} & 0.37\\
    Target context & \textit{\textsc{target} is favored to win } & 0.19 & Keyword dep. path  & \textit{\textsc{target} $\leftarrow$ will $\rightarrow$ not $\rightarrow$ \textsc{keyword}} & 0.31\\
    Keyword context & \textit{\textsc{target} are going to \textsc{keyword}} & 0.15 & Distance to keyword & \textit{\textsc{opponent} closer to \textsc{keyword}} & 0.28\\
    Target context & \textit{\textsc{target} predicted to win} & 0.13 & Target context & \textit{\textsc{target} may not win} & 0.27\\
    Pair context & \textit{\textsc{target1} could win \textsc{target2}} & 0.13 & Keyword dep. path & \textit{\textsc{opponent} $\leftarrow$ will $\rightarrow$ \textsc{keyword}} & 0.23\\
    Distance to keyword & \textit{\textsc{target} closer to \textsc{keyword}} & 0.11 & Target context & \textit{\textsc{target} can't win}& 0.18\\
    \bottomrule
    \end{tabular}}
    \caption{Some high-weight features for positive and negative veridicality.}  
    \label{tab:features_weight}
\end{table*}

\noindent
{\bf Target and opponent contexts.}
For every \textsc{target ($t$)} and \textsc{opponent ($o \in O$)} entities in the tweet, we extract context words in a window of one to four words to the left and right of the \textsc{target} (``Target context") and \textsc{opponent} (``Opponent context"), e.g., \textit{$t$ will win}, \textit{I'm going with $t$}, \textit{$o$ will win}.

\noindent
{\bf Keyword context.} For target and opponent entities, we also extract words between the entity and our specified keyword ($k$) (\textit{win} in our case): \textit{$t$ predicted to $k$}, \textit{$o$ might $k$}.

\noindent
{\bf Pair context.} For the election type of events, in which two target entities are present (contender and state. e.g., \textit{Clinton, Ohio}), we extract words between these two entities: e.g., \textit{$t_{1}$ will win $t_{2}$}.


\noindent
{\bf Distance to keyword.} We also compute the distance of \textsc{target} and \textsc{opponent} entities to the keyword. 

\noindent
\paragraph{Punctuation.} We introduce two binary features for the presence of exclamation marks and question marks in the tweet. We also have features which check whether a tweet ends with an exclamation mark, a question mark or a period. Punctuation, especially question marks, could indicate how certain authors are of their claims.

\noindent 
\paragraph{Dependency paths.} We retrieve dependency paths between the two \textsc{target} entities and between the \textsc{target} and keyword (\textit{win}) using the TweeboParser \cite{Kong14} after applying rules to normalize paths in the tree (e.g., ``doesn't" $\rightarrow$ ``does not").  

\noindent
\paragraph{Negated keyword.} We check whether the keyword is negated (e.g., ``not win", ``never win"), using the normalized dependency paths.\\

\begin{table}
    \small
    \centering
    \begin{tabular}{p{4.5cm}p{0.8cm}p{1cm}}
        \toprule
         Tweet & Gold & Predicted \\
         \midrule
         The  heart  wants  \textbf{Nadal} to  win  tomorrow  but  the  mind  points  to  a  Djokovic win  over  4  sets. Djokovic  7-5  4-6  7-5  6-4  \textbf{Nadal} for me.  & negative & positive \\ 
         \midrule
         Hopefully  tomorrow  Federer will  win  and  beat  that  \textbf{Nadal} guy lol & neutral & negative \\
         \midrule
         There  is  no  doubt \textbf{India} have  the  tools  required  to  win  their  second  World  Cup. Whether  they  do  so  will  depend on ... & positive & neutral \\
         \bottomrule         
    \end{tabular}
    \caption{Some classification errors made by TwiVer. Contenders queried for are highlighted.}
    \label{tab:error_tweets}
\end{table}

We randomly divided the annotated tweets into a training set of 
2,480 tweets, a development set of 354 tweets and a test set of 709 tweets.  MAP parameters were fit using LBFGS-B \cite{zhu1997algorithm}.
Table~\ref{tab:features_weight} provides examples of high-weight features for positive and negative veridicality.

\begin{table*}
    \small
    \centering
    \begin{tabular}{l|rrr|rrr|rrr|r}
    \toprule
    \textbf{Event} & \multicolumn{3}{c}{Veridicality}& \multicolumn{3}{c}{Sentiment} & \multicolumn{3}{c}{Frequency} &  \#predictions\\
     &P &R & F1 & P & R & F1 & P & R & F1 & \\
    \midrule
    Oscars & \textbf{80.6} & 80.6 & \textbf{80.6} & 52.9 &87.1 & 63.5 & 54.7 &\textbf{93.5} & 69.0 & 151\\
    Ballon d'Or & \textbf{100.0} & 100.0 & \textbf{100.0} & 85.7 & 100.0 & 92.2 &  85.7 & 100.0 & 92.2 & 18 \\
    Eurovision & \textbf{83.3} & 71.4 & \textbf{76.8} & 38.5 & 71.4 & 50.0  & 50.0 & 57.1 &53.3 & 87\\
    Tennis Grand Slam & 50.0 & 100.0 & 66.6 & 50.0 & 100.0 & 66.6  & 50.0 & 100.0 & 66.6 & 52\\
    Rugby World Cup & \textbf{100.0} & 100.0 & \textbf{100.0} & 50.0 & 100.0 & 66.6  & 50.0 & 100.0 & 66.6 & 4\\
    Cricket World Cup & \textbf{66.7} & 85.7 & \textbf{75.0} & 58.3 & \textbf{100.0} & 73.6 & 58.3 & \textbf{100.0} & 73.6 & 14\\
    Football World Cup & \textbf{71.4} & 100.0 & \textbf{83.3} & 62.5 & 100.0 & 76.9 & \textbf{71.4} & 100.0 & \textbf{83.3}& 10\\
    Presidential primaries &\textbf{66.0} & \textbf{88.0} & \textbf{75.4} & 58.9 & 82.5 & 68.7 & 63.4 & 78.7 & 70.2 & 211 \\
    2016 US presidential elections & 60.9 & \textbf{100.0} &\textbf{75.6}& 63.3 & 73.8 & 68.1 & \textbf{69.0} & 69.0 & 69.0 & 84 \\
    2014 Indian general elections & \textbf{95.8} & 100.0 &\textbf{97.8} & 65.6 & 91.3 & 76.3 & 56.1 & 100.0 & 71.8 & 52 \\
    \bottomrule
    \end{tabular}
    \caption{Performance of Veridicality, Sentiment baseline, and Frequency baseline on all event categories (\%).}
    \label{tab:system_comparison}
\end{table*}

\subsection{Evaluation}
We evaluated TwiVer's precision and recall on our held-out test set of 709 tweets.
Figure~\ref{fig:pos_veridic} shows the precision/recall curve for positive veridicality.
By setting a threshold on the probability score to be greater than $0.64$, we achieve a precision of $80.1\%$ and a recall of $44.3\%$ in identifying tweets expressing a positive veridicality toward a candidate winning a contest.

\subsection{Performance on held-out event types}
To assess the robustness of the veridicality classifier when applied to new types of events, we compared its performance when trained on all events vs.\ holding out one category for testing. Table~\ref{tab:cross_domain} shows the comparison: the second and third columns give F1 score when training on all events vs. removing tweets related to the category we are testing on. In most cases we see a relatively modest drop in performance after holding out training data from the target event category, with the exception of elections.  This suggests our approach can be applied to new event types without requiring in-domain training data for the veridicality classifier.




\begin{table*}[h]
    \small
    \centering
    \begin{tabular}{lrrr}
    \toprule
    Event & Train on all & Train without held-out event & $|T_t|$ \\
    \midrule
    Oscars & 69.5 & 63.8 & 64\\ 
    Ballon d'Or & 54.6 & 46.6 & 61 \\  
    Eurovision & 65.7 & 63.2 & 48 \\  
    Tennis Grand Slams & 52.1 & 45.5 & 44 \\  
    Rugby World Cup & 56.5 & 58.1 & 44 \\  
    Cricket World Cup & 61.9 & 66.8 & 49 \\ 
    Football World Cup & 76.0 & 67.5 & 56 \\ 
    Presidential primaries & 59.8 & 48.1 & 117 \\ 
    2016 US presidential elections & 52.0 & 52.3 & 54 \\ 
    Indian elections & 60.3 & 39.0 & 44 \\ 
    
    \bottomrule
    \end{tabular}
    \caption{F1 scores for each event when training on all events vs.\ holding out that event from training. $|T_t|$ is the number of tweets of that event category present in the test dataset.}
    \label{tab:cross_domain}
\end{table*}



\subsection{Error Analysis}

Table~\ref{tab:error_tweets} shows some examples which TwiVer incorrectly classifies. These errors indicate that even though shallow features and dependency paths do a decent job at predicting veridicality, deeper text understanding is needed for some cases. The opposition between ``the heart \ldots the mind" in the first example is not trivial to capture. Paying attention to matrix clauses might be important too (as shown in the last tweet ``There is no doubt \ldots"). 

\section{Forecasting Contest Outcomes}
\label{forecasting_contest_outcomes}
We now have access to a classifier that can automatically detect positive veridicality predictions about a candidate winning a contest.  This enables us to evaluate the accuracy of the crowd's wisdom by retrospectively comparing popular beliefs (as extracted and aggregated by TwiVer) against known outcomes of contests.

We will do this for each award category (Best Actor, Best Actress, Best Film and Best Director) in the Oscars from 2009 -- 2016, for every state for both Republican and Democratic parties in the 2016 US primaries, for both the candidates in every state for the final 2016 US presidential elections, for every country in the finals of Eurovision song contest, for every contender for the Ballon d'Or award, for every party in every state for the 2014 Indian general elections, and for the contenders in the finals for all sporting events. 



\begin{table*}[t]
    \small
    \centering
    \begin{tabular}{llllp{0.2cm}ll}
    \toprule
     & \multicolumn{3}{c}{Veridicality} & &  \multicolumn{2}{c}{Sentiment}\\
    && Contender &     Score & &  Contender &  Score\\
    \midrule
    \textsc{oscars} && \textbf{Leonardo DiCaprio}  & 		0.97 &&\textbf{Julianne Moore} &  0.85 \\
    && \textbf{Natalie Portman} & 	0.92 && Mickey Rourke &  0.83\\
    && \textbf{Julianne Moore}  & 	 0.91 && \textbf{Leonardo DiCaprio (2016)} &   0.82 \\
    && \textbf{Daniel Day-Lewis} &		0.90 && \textbf{Kate Winslet} &   0.75 \\
    && \textbf{Slumdog Millionaire} &	0.75 & & Leonardo DiCaprio (2014) & 	0.69 \\
    && \textbf{Matthew McConaughey} & 	0.74 && \textbf{Slumdog Millionaire}  & 0.67 \\
    &{\bf !} & The Revenant &  0.73 && \textbf{Danny Boyle} & 	0.67 \\
    && \textbf{Argo} &	0.71 && \textbf{Daniel Day-Lewis} & 	0.66 \\
    && \textbf{Brie Larson} &	0.70 && \textbf{Brie Larson} &	 0.65 \\
    && \textbf{The Artist} & 	0.67 && George Miller &	0.63 \\
   \midrule
    \textsc{primaries}  && \textbf{Trump} \hspace*{0.3cm} 	South Carolina  & 0.96 && \textbf{Sanders} \hspace*{0.18cm}	West Virginia &	0.96\\
    && \textbf{Clinton} \hspace*{0.21cm}  Iowa  &  0.90 && \textbf{Clinton} \hspace*{0.23cm}	 North Carolina	 &	0.93\\
    && \textbf{Trump} \hspace*{0.3cm} Massachusetts  & 0.88 && \textbf{Trump} \hspace*{0.31cm}	North Carolina  & 0.91\\
    && \textbf{Trump} \hspace*{0.3cm} Tennessee & 0.88 && \textbf{Sanders} \hspace*{0.18cm}	Wyoming  &	0.90\\
   && \textbf{Sanders} \hspace*{0.18cm} 	Maine &	0.87 && \textbf{Sanders} \hspace*{0.18cm}	Oklahoma  &	0.89\\
    && \textbf{Sanders} \hspace*{0.18cm}	Alaska  &	0.87 && \textbf{Sanders} \hspace*{0.18cm}	Hawaii  & 0.86\\
    & {\bf !} & Trump \hspace*{0.37cm}	Maine  &	0.87 && Sanders  \hspace*{0.22cm}	Arizona  &	0.86\\
    && \textbf{Sanders} \hspace*{0.18cm}	Wyoming  & 0.86 && \textbf{Sanders} \hspace*{0.18cm}	Maine  &	0.85\\
    && \textbf{Trump} \hspace*{0.32cm}	Louisiana  & 0.86 && \textbf{Trump}  \hspace*{0.3cm}	Delaware  &	0.84\\
    & {\bf !} & Clinton \hspace*{0.3cm}	Indiana  &	0.85 && \textbf{Trump}  \hspace*{0.3cm}	West Virginia  &	0.83\\    
    \bottomrule
    \end{tabular}
    \caption{Top 10 predictions of winners for Oscars and primaries based on veridicality and sentiment scores. Correct predictions are highlighted. ``{\bf !}" indicates a loss which wasn't expected.}
    \label{tab:veridicality_prediction}
\end{table*}

\subsection{Prediction}
\label{prediction}

A simple voting mechanism is used to predict contest outcomes: we collect tweets about each contender written before the date of the event,\footnote{These are a different set of tweets than those TwiVer was trained on.} and use TwiVer to measure the veridicality of users' predictions toward the events. Then, for each contender, we count the number of tweets that are labeled as positive with a confidence above 0.64, as well as the number of tweets with positive veridicality for all other contenders. Table~\ref{tab:oscar_actress} illustrates these counts for one contest, the Oscars Best Actress in 2014.

\begin{table}
    \small
    \centering
    \begin{tabular}{lrr}
    \toprule
    Contender & $|T_c|$ & $|T_O|$ \\ 
    \midrule
    \textbf{Cate Blanchett} & 73 & 46 \\ 
    Amy Adams & 6 & 113 \\ 
    Sandra Bullock & 22 & 97 \\ 
    Judi Dench & 2 & 117 \\ 
    Meryl Streep & 16 & 103 \\ 
    \bottomrule
    \end{tabular}
    \caption{Positive veridicality tweet counts for the Best Actress category in 2014: $|T_c|$ is the count of positive veridicality tweets for the contender under consideration and $|T_O|$ is the count of positive veridicality tweets for the other contenders.}
    \label{tab:oscar_actress}
\end{table}
We then compute a simple prediction score, as follows:
\begin{equation}
\text{score} = (|T_c| + 1)/(|T_c| + |T_O| + 2)    
\label{eq:score}
\end{equation}
where $|T_c|$ is the set of tweets mentioning positive veridicality predictions toward candidate $c$, and $|T_O|$ is the set of all tweets predicting any opponent will win.
For each contest, we simply predict as winner the contender whose score is highest.

\subsection{Sentiment Baseline}
We compare the performance of our approach against a state-of-the-art sentiment baseline \cite{Saif13}. Prior work on social media analysis used sentiment to make predictions about real-world outcomes. For instance, \citet{o2010tweets} correlated sentiment with public opinion polls and
\citet{GermanElections} use political sentiment to make predictions about outcomes in German elections.

We use a re-implementation of \cite{Saif13}'s system\footnote{\url{https://github.com/ntietz/tweetment}} to estimate sentiment for tweets in our corpus. We run the tweets obtained for every contender through the sentiment analysis system to obtain a count of positive labels. Sentiment scores are computed analogously to veridicality using Equation \eqref{eq:score}. For each contest, the contender with the highest sentiment prediction score is predicted as the winner.

\subsection{Frequency Baseline}
We also compare our approach against a simple frequency (tweet volume) baseline. For every contender, we compute the number of tweets that has been retrieved. Frequency scores are computed in the same way as for veridicality and sentiment using Equation \eqref{eq:score}. For every contest, the contender with the highest frequency score is selected to be the winner.

\subsection{Results}

Table~\ref{tab:system_comparison} gives the precision, recall and max-F1 scores for veridicality, sentiment and volume-based forecasts on all the contests. The veridicality-based approach outperforms sentiment and volume-based approaches on 9 of the 10 events considered. For the Tennis Grand Slam, the three approaches perform poorly. The difference in performance for the veridicality approach is quite lower for the Tennis events than for the other events. It is well known however that winners of tennis tournaments are very hard to predict. The performance of the players in the last minutes of the match are decisive, and even professionals have a difficult time predicting tennis winners.

Table~\ref{tab:veridicality_prediction} shows the 10 top predictions made by the veridicality and sentiment-based systems on two of the events we considered - the Oscars and the presidential primaries, highlighting correct predictions.

\subsection{Surprise Outcomes}
\label{surprise}
In addition to providing a general method for forecasting contest outcomes, our approach based on veridicality allows us to perform several novel analyses including retrospectively identifying surprise outcomes that were unexpected according to popular beliefs.

In Table~\ref{tab:veridicality_prediction}, we see that the veridicality-based approach incorrectly predicts \textit{The Revenant} as winning Best Film in 2016.  This makes sense, because the film was widely expected to win at the time, according to popular belief. Numerous sources in the press,\footnote{www.forbes.com/sites/zackomalleygreenburg/2016/02/29/
spotlight-best-picture-oscar-is-surprise-of-the-night/\#52f546c2721a}\footnote{www.vox.com/2016/2/26/11115788/revenant-best-picture}\footnote{www.mirror.co.uk/tv/tv-news/spotlight-wins-best-picture-2016-7460633} qualify \textit{The Revenant} not winning an Oscar as a big surprise.

Similarly, for the primaries, the two incorrect predictions made by the veridicality-based approach were surprise losses. 
News articles\footnote{http://patch.com/us/across-america/maine-republican-caucus-live-results-trump-favored-win-0}\footnote{http://www.huffingtonpost.com/entry/ted-cruz-upset-win-maine-republican-caucus\_us\_56db461ee4b0ffe6f8e9a865}\footnote{https://news.vice.com/article/bernie-sanders-wins-indiana-primary-in-surprise-upset-over-hillary-clinton} indeed reported the loss of Maine for Trump and the loss of Indiana for Clinton as unexpected.


\subsection{Assessing the Reliability of Accounts}
\label{assessing}
Another nice feature of our approach based on veridicality is that it immediately provides an intuitive assessment on the reliability of individual Twitter accounts' predictions. For a given account, we can collect tweets about past contests, and extract those which exhibit positive veridicality toward the outcome, then simply count how often the accounts were correct in their predictions.

As proof of concept, we retrieved within our dataset, the user names of accounts whose tweets about Ballon d'Or contests were classified as having positive veridicality. 
Table~\ref{tab:oscar_predictions} gives accounts that made the largest number of correct predictions for Ballon d'Or awards between 2010 to 2016, sorted by users' prediction accuracy. Usernames of non-public figures are anonymized (as user 1, etc.) in the table.  We did not extract more data for these users: we only look at the data we had already retrieved. Some users might not make predictions for all contests, which span 7 years.

\begin{table}
    \small
    \centering
    \resizebox{0.5\textwidth}{!}{
    \begin{tabular}{lr|lr}
    \toprule
    User account & Accuracy(\%) & User account & Accuracy(\%)\\
    \midrule
     User 1   & 100 (out of 6)  & twitreporting  & 100 (out of 3)\\
     Cr7Prince4ever & 100 (out of 6) &  User 3 & 100 (out of 3)\\
     goal\_ghana & 100 (out of 4) & Naijawhatsup & 100 (out of 3)\\
     User 2 & 100 (out of 4) & 1Mrfutball & 90 (out of 10)\\
     breakingnewsnig & 100 (out of 4) & User 4 & 77 (out of 13)\\
    \bottomrule
    \end{tabular}}
    \caption{List of users sorted by how accurate they were in their Ballon d'Or predictions.}
    \label{tab:oscar_predictions}
\end{table}

Accounts like ``goal\_ghana", ``breakingnewsnig" and ``1Mrfutball", which are automatically identified by our analysis, are known to post tweets predominantly about soccer.




\section{Conclusions}
In this paper, we presented TwiVer, a veridicality classifier for tweets which is able to ascertain the degree of veridicality toward future contests.  We showed that veridical statements on Twitter provide a strong predictive signal for winners on different types of events, and that our veridicality-based approach outperforms a sentiment and frequency baseline for predicting winners. 
Furthermore, our approach is able to retrospectively identify surprise outcomes. We also showed how our approach enables an intuitive yet novel method for evaluating the reliability of information sources.

\section*{Acknowledgments}
We thank our anonymous reviewers for their valuable feedback. We also thank Wei Xu, Brendan O'Connor and the Clippers group at The Ohio State University for useful suggestions.
This material is based upon work supported by the National Science Foundation under Grants No.\ IIS-1464128 to Alan Ritter and IIS-1464252 to Marie-Catherine de Marneffe. Alan Ritter is supported by the Department of Defense under Contract No. FA8702-15-D-0002 with Carnegie Mellon University for the operation of the Software Engineering Institute, a federally funded research and development center in addition to the Office of the Director of National Intelligence (ODNI) and the Intelligence Advanced Research Projects Activity (IARPA) via the Air Force Research Laboratory (AFRL) contract number FA8750-16-C-0114. The U.S.\ Government is authorized to reproduce and distribute reprints for Governmental purposes notwithstanding any copyright annotation thereon. The views and conclusions contained herein are those of the authors and should not be interpreted as necessarily representing the official policies or endorsements, either expressed or implied, of ODNI, IARPA, AFRL, NSF, or the U.S.\ Government.


\bibliography{emnlp2017}
\bibliographystyle{emnlp_natbib}

\end{document}